\documentclass[conference]{IEEEtran}
\usepackage[T1]{fontenc}
\usepackage{amsmath,graphicx,algorithmic,algorithm2e}
\usepackage[noadjust]{cite}
\usepackage{epstopdf}
\usepackage{subfigure}
\usepackage{gensymb}
\usepackage{amssymb}

\usepackage{etoolbox}
\makeatletter
\patchcmd{\@makecaption}
  {\scshape}
  {}
  {}
  {}
\makeatother
\def\tablename{Table}

\begin{document}
% Title.
% ------
\title{Incremental Learning of Environment Interactive Structures from Trajectories of Individuals}
%
% Single address.
% ---------------
\author{\IEEEauthorblockN{Damian Campo\IEEEauthorrefmark{1},
Vahid Bastani,
Lucio Marcenaro and 
Carlo Regazzoni}
\IEEEauthorblockA{Department of Naval, Electric, Electronic and Telecommunications Engineering\\
University of Genova, Italy\\
 \IEEEauthorrefmark{1}Email: damian.campo@ginevra.dibe.unige.it}}

%
% For example:
% ------------
%\address{School\\
%	Department\\
%	Address}
%
% Two addresses (uncomment and modify for two-address case).
% ----------------------------------------------------------
%\twoauthors
%  {A. Author-one, B. Author-two\sthanks{Thanks to XYZ agency for funding.}}
%	{School A-B\\
%	Department A-B\\
%	Address A-B}
%  {C. Author-three, D. Author-four\sthanks{The fourth author performed the work
%	while at ...}}
%	{School C-D\\
%	Department C-D\\
%	Address C-D}
%

%\ninept
%
\maketitle
\begin{abstract}
This work proposes a novel method for estimating the influence that unknown static objects might have over mobile agents. Since the motion of agents can be affected by the presence of fixed objects, it is possible use the information about trajectories deviations to infer the presence of obstacles and estimate the forces involved in a scene. Artificial neural networks are used to estimate a non-parametric function related to the velocity field influencing moving agents. The proposed method is able to incrementally learn the velocity fields due to external static objects within the monitored environment. It determines whether an object has a repulsive or an attractive influence and provides an estimation of its position and size. As stationarity is assumed, i.e., time-invariance of force fields, learned observation models can be used as prior knowledge for estimating hierarchically the properties of new objects in a scene.
\end{abstract}
%
%\begin{keywords}
%Kalman Filtering, Force Interaction Models, Artificial Neural Networks, Situation Awareness.
%\end{keywords}
%

\section{Introduction}
\label{sec:intro}

Analysis of agents whose motion follows some patterns or rules is a topic that has been of increasing interest in recent years. Mobile agent behaviors are frequently analyzed in multiple research fields, e.g., video surveillance, crowd monitoring, sport events and situation awareness \cite{Hsieh2008, Cristani2013, Piriou2006}. This paper aims towards the representation and incremental learning of causal motivations modifying the state of moving agents in an observable way. Incremental learning is able to deal with situations where new static objects (attractors or repellers) are progressively added. Observations that produce deviations from previous knowledge are used to increase the environmental understanding. Accordingly, new models can be characterized by comparing expected learned behaviors with the observed one.

Cognitive systems that analyze data from unstructured information represent a fundamental topic in which a wide research is under development \cite{Aggarwal2011}. Due to the large amount of information that can be collected from technological devices such as positioning and locating systems, it becomes important to provide methods that allow not only to detect and track objects robustly but also to extract meaningful information to explain the contextual motivation that can be related to their motion. 

The Bayesian inference methodology has proven to be a powerful tool to model and explain interactive dynamical behaviors through a probabilistic framework. Several problems can be solved using a stochastic formulation, e.g., crowd monitoring \cite{Napoletano2015,Calderara2008}, crowd analysis \cite{ Ferrer2014,Wu2014,Andrade2006,Chiappino2015}, event detection \cite{Wu2014,Andrade2006,Hongeng2004,Dore2010}, etc. 

Understanding the context of video sequences is an important task that improves the identification of events involving interactions among agents in a scene. Detection of events from video is a topic that is gaining interest in the research community. Several works have been presented in this area, some of them are focused on detection of abnormal behaviors in surveillance videos \cite{Mehran2009,Yuan2015,Chiappino2015,Vahid2016,1020947}, detection of groups of people in scenes \cite{Chan2012,Vahid2015} and identification of dangerous massive crowds  \cite{Bu2011}. This particular work is focused on cases in which no visual information is available but the system can acquire and process data about the dynamic localization of the involved agents.  

When a particular object is introduced in the scene, it becomes essential to know the effect that it exerts on the state of moving agents that interact with it. From that point of view, Situation Awareness (SA) can be seen as a problem of joint data estimation where a set of measurements about the position of agents provide information about the motivation that makes them move in a certain way.

Particular zones of the environment can be considered as repulsive or attractive static objects creating a force field over agents, that causes a modification of their motion. The knowledge of a learned force field can be used to express other environmental constraints by using the agent deviation from the expected model already characterized.

The core of this work relies on the understanding of interactions between mobile agents and static objects. By observing the motion of agents it is possible to infer characteristics of external objects through the estimation of velocity fields associated with their effect on the environment. A Bayesian method has been selected for this task due to its capacity of considering probabilities to model interdependent events with their related uncertainty. Bayesian Models or Probabilistic Graphical Models (PGMs) \cite{Koller2009} are able to represent and temporally predict upcoming situations and they have proved to be useful in SA problems \cite{Park2014,Costa2009,Carvalho2010}.

Formally, SA is defined as ``the perception of the elements within a volume of time and space, the comprehension of their meaning, the explanation of their present (observed) status and the ability to project the same in the near future'' \cite{Bhatt2012}. From this perspective, SA represents a relevant issue to consider for understanding influences of objects inside scenes. The goal of this work is to identify repulsive and attractive properties of unknown objects given the movement of agents in the scene. 

The paper is organized as follows: Section \ref{sec:Problem} gives a brief introduction to the concepts of force fields, section \ref{sec:ProblemDes} describes the problem that will be tackled, section \ref{sec:format} presents the proposed method, while results are analyzed in section \ref{sec:results} and conclusions are drawn in section \ref{sec:typestyle}. 

\section{Force field terminology}
\label{sec:Problem}

Taking into consideration a classical mechanics approach, a force is defined as a vectorial quantity that acts on a body to cause a change in its state of motion \cite{Wagh2012}. Forces can be classified in action-reaction (when bodies, which are in contact, change their momenta \cite{Wagh2012}) and action-at-a-distance forces (when objects interact without being physically touched).

Considering that social interactions can be often modeled as contact-less, it becomes possible to explain social phenomena in a certain environment by modeling interactions between entities with action-at-a-distance forces. 

A force field $\vec{F}$ is defined as a vector point-function which has the property that at every point of the space takes a particular value related to the magnitude and direction of a force acting on a particle of unit of mass placed there \cite{Tenenbaum2012}. Accordingly, in this work, the particles of unit of mass affected by force fields will be called agents. 

A central force field $\vec{F}=f(r)\hat{r}$ is a special case of force field in which the motion of agents is affected depending on the distance  $r$ to a center of force, which is generally associated with the center of mass of the object that produces the force field. $\hat{r}$ is a unit vector in the direction of $r$.

Additionally, a force field is called conservative, if it can be expressed in terms of the gradient of a function $\Phi(x,y,z)$. In the case of central force fields, since they are spherically symmetric, they will be conservative as well, such that:

\begin{equation} \label{eq3c}
\vec{F}=f(r)\hat{r}=-\nabla \Phi(r)
\end{equation}

As an example, let $\Phi_{g}(r)$ be the potential function of the gravitational force field, such that:

\begin{equation} \label{eq3e}
\Phi_{g}(r)=-\frac{GMm}{r}
\end{equation}

Where m and M are respectively the masses of the agent and the object that exerts the force field. $G  = 6.67 \times 10^{-11} \frac {N m^{2}}{kg^{2}} $ is the gravitational constant and $r$ is the distance between the agent and the object's center of mass.  

By substituting the potential function shown in equation (\ref{eq3e}) into equation (\ref{eq3c}), it is possible to obtain the gravitational force $\vec{F}_{g}$ in (\ref{eq3f}).

\begin{equation} \label{eq3f}
\vec{F}_{g}=-\frac{GMm}{r^{2}}\hat{r}
\end{equation}

\begin{figure}[!htb]
  \begin{minipage}[b]{0.48\columnwidth}  
    \centering
    \includegraphics[width=1.05\linewidth]{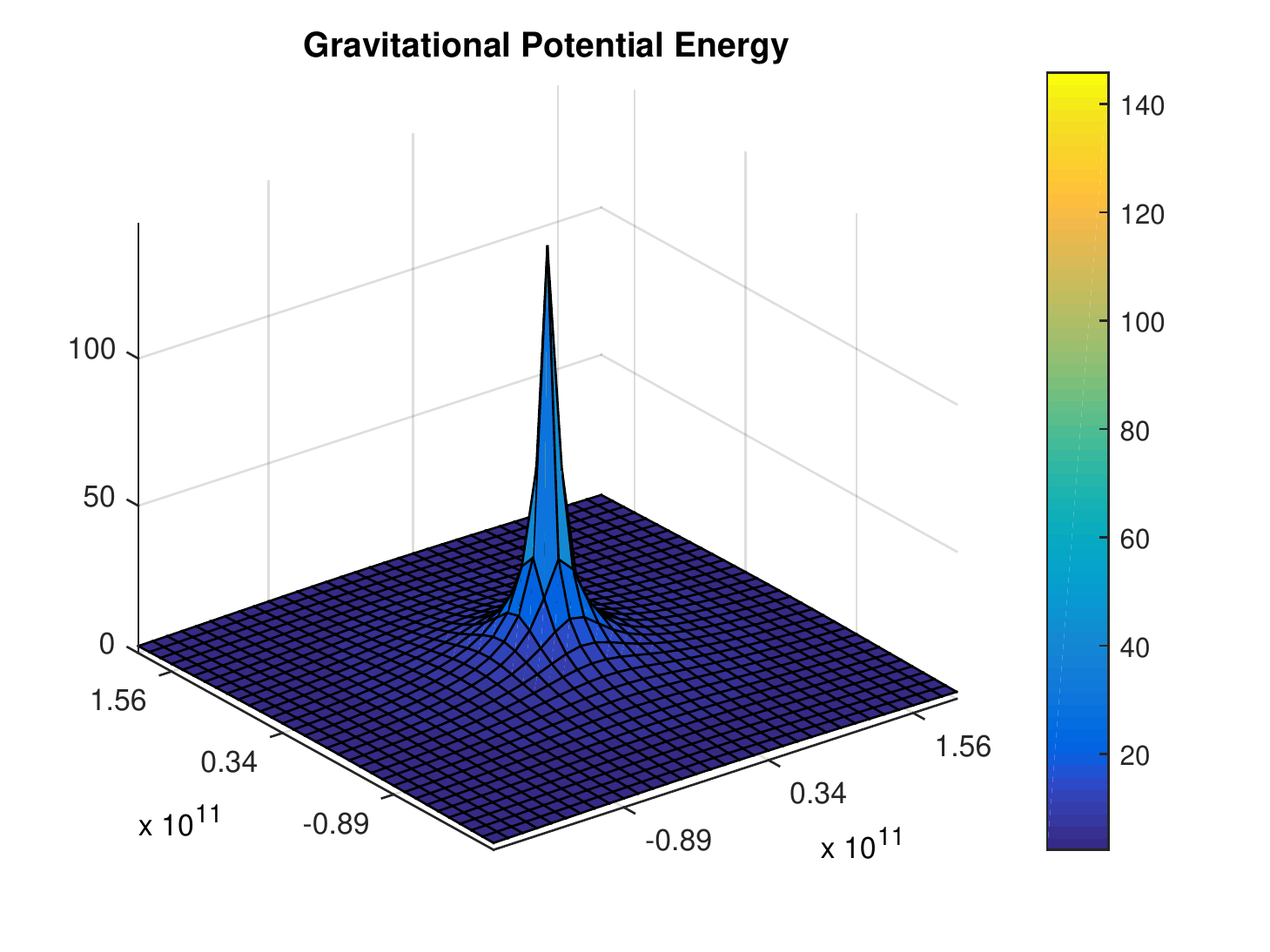} 
    \caption{Gravitational Potential Energy} 
     \label{fig:1a}
    \vspace{1ex}
  \end{minipage}%%
\hfill
  \begin{minipage}[b]{0.48\columnwidth}
    \centering
    \includegraphics[width=1.05\linewidth]{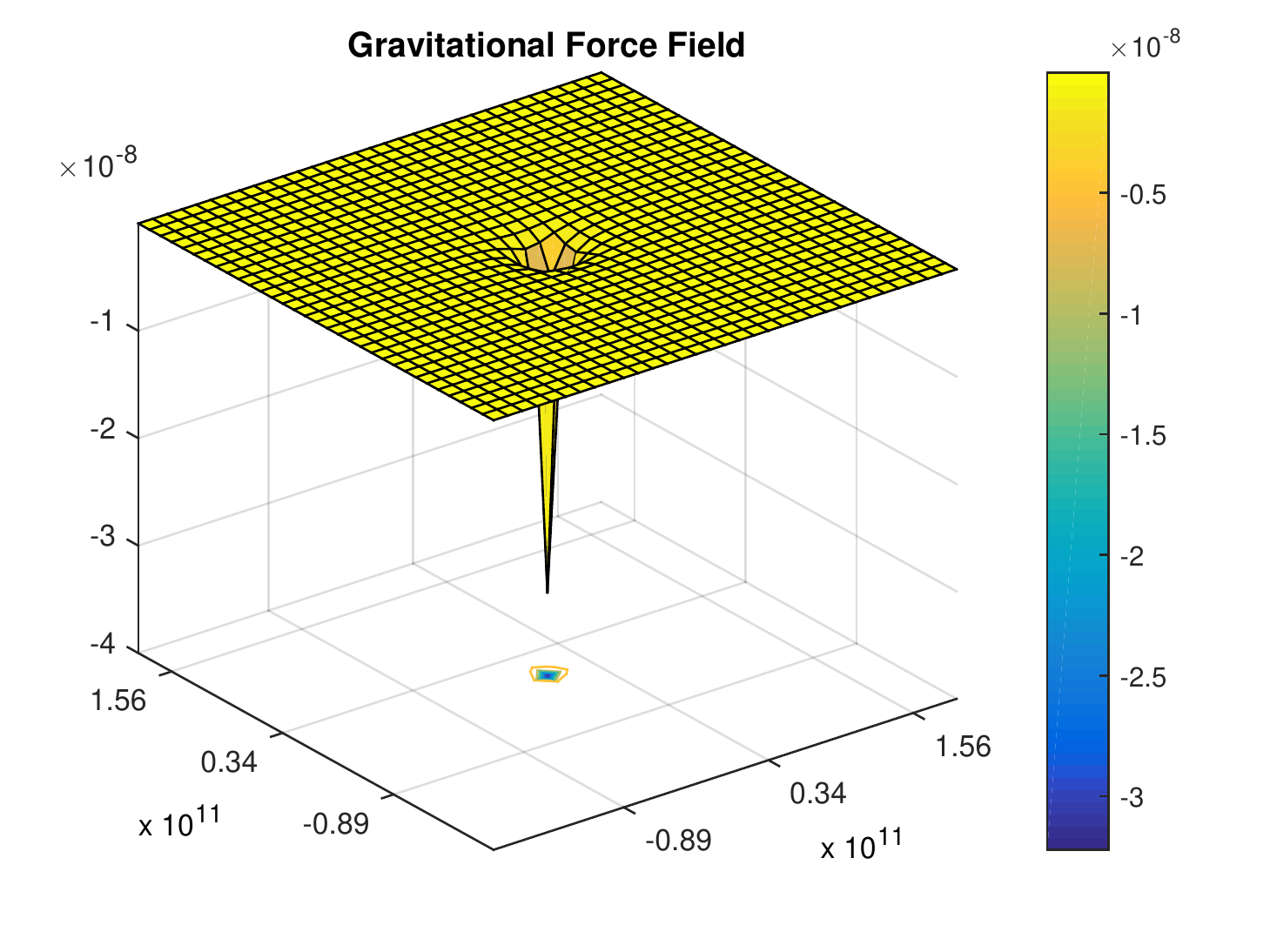} 
    \caption{Gravitational Force Field} 
    \label{fig:1b}
    \vspace{3ex}
  \end{minipage} 
\hfill
  \begin{minipage}[b]{1\columnwidth}
    \centering
    \includegraphics[width=0.65\linewidth]{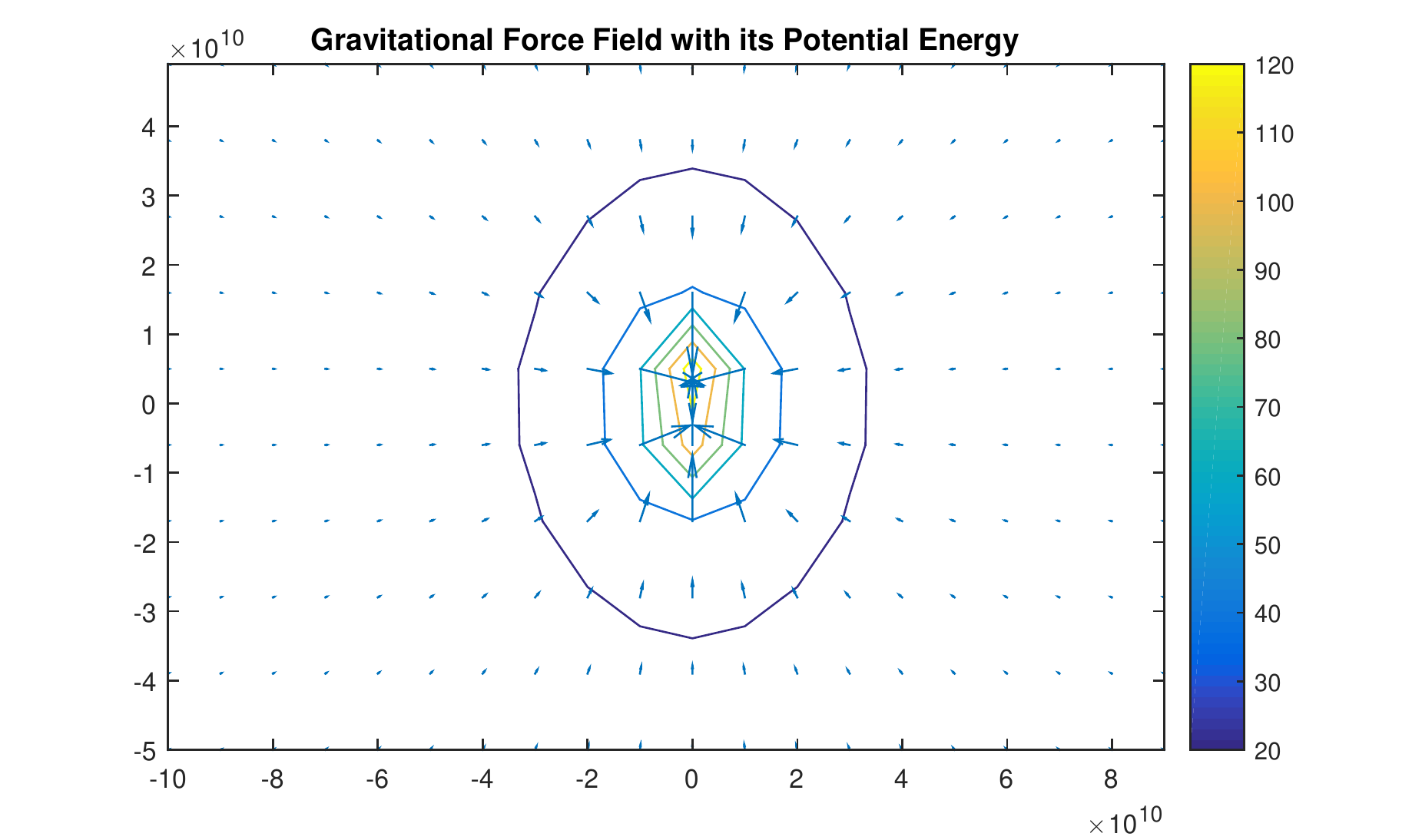} 
    \caption{Gravitational Force Field with its Potential Energy} 
    \label{fig:1c}
    \vspace{3ex}
  \end{minipage} 
\end{figure}

In Fig. \ref{fig:1a} and Fig. \ref{fig:1b} are presented the plots of the potential energy $-\Phi_{g}(r)$ and the force field $\vec{F}_{g}$ respectively. For visualization purposes, in both figures a two-dimensional plane whose units are given in meters was considered in the axes $x$ and $y$. The center of force is placed in the origin of such plane i.e., $(0,0)$ and both masses, M and m, are considered as $1kg$. 

In Fig. \ref{fig:1a}, the $z$ axis shows the magnitude of the gravitational potential. Fig. \ref{fig:1b} $z$ represents the magnitude of the force field. The effect of both is plotted in Fig. \ref{fig:1c}. 

Accordingly, this paper is focused on characterizing the effects of central force fields $f(r)\hat{r}$ in terms of agent motions. 

\section{Problem description}
\label{sec:ProblemDes}

\subsection{Continuous Case}
\label{sssec:Continuous}

The dynamic equation for describing the kinematic motion of an agent with constant acceleration is described below:

\begin{equation} \label{eq1aa}
p(t +\Delta t)= p(t) + v(t)\Delta t + \frac{a \Delta t^2 }{2}
\end{equation}

Where $p(t)$ represents the agent position in a certain instant $t$, $v(t)$ is its velocity and $a$ its acceleration. By substituting the expression of force $F=ma$ in equation (\ref{eq1aa}), it is possible to express the force in terms of the agent motion, such that: 

\begin{equation} \label{eq1ab}
F=2m\bigg[\frac{p(t +\Delta t)- p(t) - v(t)\Delta t}{\Delta t^2}\bigg]
\end{equation}

Considering that the current and next position of the agent i.e., $p(t)$ and $p(t +\Delta t)$ are known, it is possible to obtain the force $F$ by calculating the parameter $v(t)$. Taking advantage from the additive property of velocities, it is possible to express such parameter as follows:

\begin{equation} \label{eq1ac}
v(t)=\sum_i^N v^i(t)
\end{equation} 

Where $v^i(t)$ represents the velocity due to a force field produced by an external object $i$.

The main idea of this work is to incrementally learn the velocities $v^i(t)$ associated with the force fields produced by static objects. If such velocities become space dependent, it is possible to define a velocity field $v(\mathbb{R}^2)$ for a two dimensional environment. The description of the problem in terms of discrete time is formulated in the following. 

\subsection{Discrete Case}
\label{sssec:Discrete}

By observing trajectories in a two-dimensional space $\mathbb{R}^2$, it is possible to define $Z_k$ as a vector of observations of the agent position $(x,y)$ at a time instant $k$. The state of the agent $X_{k}$ is defined as its position and velocity at a time $k$.

The relation between measurements of agent positions $Z_k$ and its state $X_{k}$ is expressed in equation (\ref{eq5}). 

\begin{equation} \label{eq5}
Z_{k} = HX_{k} + \nu_{k}
\end{equation}

Where, 

$$
 X_k=\left[ \begin{array}{c} x_k \\ y_k \\ \dot{ x_k}\\ \dot{ y_k} \end{array} \right]  ; H=\begin{bmatrix} 1 & 0  &  0 & 0\\ 0 & 1  & 0 & 0 \end{bmatrix} 
$$

$\nu_{k}$ represents the noise in an instant $k$ produced by the sensor that is measuring the position of the agent. 
 
It is possible to consider the next general dynamic model for describing the motions of agents:  

\begin{equation} \label{eq4aa}
X_{k+\Delta k} = FX_{k} + BV_{k} + n_{k}
\end{equation}

Where, 

$$
 F=\begin{bmatrix} 1 & 0  &  0 & 0\\ 0 & 1  & 0 & 0 \\ 0 & 0  & 0 & 0 \\ 0 & 0  & 0 & 0\end{bmatrix} ;  B=\begin{bmatrix} \Delta k& 0  \\ 0 & \Delta k  \\ 1 & 0  \\ 0 & 1 \end{bmatrix} ; V_k=\left[ \begin{array}{c}   \dot{ x_k} \\ \dot{ y_k} \end{array} \right] 
$$

Where $V_{k}$ is the velocity of the agent that is due to the presence of fixed objects inside the scene at time $k$; parameter $n_k$ represents the noise of the dynamic model at time $k$. 

If $V_k$ is considered as the sum of all the effects produced by the objects interacting with the agent at a time $k$, it is possible to rewrite the term $V_k$, such that:

\begin{equation} \label{eq4ab}
 V_{k}= \sum_{i=1}^{N} V^{i}_{k}
\end{equation}

Where $N$ represents the total number of objects that affect the state of the agent. From equation (\ref{eq4ab}), it is possible to see that the classic mechanics additivity of velocities is used to include all the external influences into the proposed dynamic formulation that estimates the state of agents in time. 

The sum of velocities caused by different objects inside an environment allows a hierarchical representation of the scene. The effect of each object can be incrementally learned by comparing the current motion of an agent with models that were learned from previous observations. Section \ref{sec:format} explains the method that is proposed to do so.

\section{Proposed Method}
\label{sec:format}

Let's start considering the simplest dynamical model where null velocity contributions are assumed, i.e., a non-interactive model. Its formulation is shown in equation (\ref{eq4ac}).

\begin{equation} \label{eq4ac}
X_{k+\Delta k}^1= F{X}_{k}^1 + n_k
\end{equation}

Where $X_{k+\Delta k}^1$ is the estimation of the next point based on a non-interactive dynamical model. In this way, deviations from such a model will give information about the effects of unknown objects in the scene. 

By taking an observation at time instant $k+\Delta k$, it is possible to obtain the term $\tilde{Y}^1_{k}$, associated with the difference between the state estimation obtained from a non-interactive dynamical model and the measurement taken by a sensor, such that: 

\begin{equation} \label{eq4ad}
\tilde{Y}_{k}^1 = Z_{k+\Delta k} - H X_{k+\Delta k}^1
\end{equation}

$\tilde{Y}_{k}^1$ is the innovation or residual measurement produced by the non-interactive model. In the ideal case, $\tilde{Y}_{k}^1$ goes to zero, that means the proposed dynamical model correctly describes the behavior of the agent.

Given the case in which $\tilde{Y}_{k}^1$ is significantly different from zero, it is necessary to propose a more complete model that uses the residual of the non-interactive model $\tilde{Y}_{k}^1$ to accurately predict the agent behavior. From this perspective, it is considered the next new dynamical model:

\begin{equation} \label{eq4aab0}
{X}_{k+\Delta k}^2= {X}_{k+\Delta k}^1 + BV_k^1
\end{equation} 
The motivation of the above expression relies on the fact of modeling the effect of unknown objects as a velocity term $BV_k^1$, see equation (\ref{eq4aa}).
Accordingly, the innovation of the new model, $\tilde{Y}_{k}^2$, must be close to zero, such that:

\begin{equation} \label{eq4aab}
Z_{k+\Delta k} - H {X}_{k+\Delta k}^2=0
\end{equation}  

By considering equations (\ref{eq4ad}) and (\ref{eq4aab0}) in the expression of equation (\ref{eq4aab}), it is possible to obtain $V_k^1$ in terms of $\tilde{Y}_{k}^1$, such that:

\begin{equation} \label{eq4ae}
\tilde{Y}_{k}^1 = H B {V}_{k}^1  \Longrightarrow V_k^1=\frac{\tilde{Y}^1_{k}}{\Delta k}
\end{equation}

In this sense, a more complete dynamic model, which includes more knowledge about forces involved in the environment, can be used as a new estimator as it is shown below. 

\begin{equation} \label{eq4aaa}
X_{k+\Delta k}^2 = FX_{k}^2+ B\frac{\tilde{Y}_{k}^1}{\Delta k} + n_{k}
\end{equation}

One of the purposes of the present work is to learn the parameters $\frac{\tilde{Y}_{k}^i}{\Delta k}$ related to each static object $i$. 

Given the case that an object $i$ appears in the scene, it is possible to extract information about its effect by calculating $\frac{\tilde{Y}_{k}^i}{\Delta k}$. This process is done by using the non-interactive model, see equation (\ref{eq4ac}), together with the $i-1$ models previously estimated.

The non-interactive model can be seen as a baseline from which other models that include information about static objects can be created. 

Through the characterization of the effects of several objects in the scene, the knowledge about the environment is increased and more complete dynamic models that describe the motion of agents can be obtained. 

In general, supposing that $N$ static objects are placed in the environment one by one, such that it is possible to characterize each of them from the knowledge previously acquired, it is possible to write the next general expression for the hierarchical formulation of models:

\begin{equation} \label{eq4aad}
X_{k+\Delta k}^{N+1} = FX_{k}^{N+1} + B\sum_{i=0}^{N} \frac{\tilde{Y}^{i}_{k}}{\Delta k} + n_{k}
\end{equation}

Where $\tilde{Y}_k^{0} = 0$, which represents a non-interactive model.

An equivalent expression of equation (\ref {eq4ab}), that expresses the sum of $N$ objects' effects in terms of agent velocities added in a hierarchical way, is defined as follows:

\begin{equation} \label{eq4aac}
 V_{k}= \sum_{i=0}^{N} \frac{\tilde{Y}^{i}_{k}}{\Delta k}
\end{equation}

Where $\tilde{Y}^i_{k}$ is the innovation or residual generated by the object $i$. As discussed before, this term is defined as the difference between observed agent position and dynamical model prediction, see equation (\ref{eq4ad}). $\Delta k$ represents the sample time. 

By supposing that time invariant force fields are caused by static objects, it is possible to make $\tilde{Y}_{k}$ dependent on the agent location only, i.e., on the measured position $Z$:

\begin{equation} \label{eq4af}
\tilde{Y}_{k} = \tilde{Y}_{Z}  \Longrightarrow V_k=V_Z
\end{equation}

Where $\tilde{Y}_{Z}$ and $V_Z$ are spatially dependent. Accordingly, considering an unknown object $i$ that produces a central force field $f(r_{i})\hat{r_{i}}$ and therefore a velocity field $\upsilon(r_{i})\hat{r_{i}}$, it becomes necessary to calculate the parameter $r_{i}$ to approximate both fields. Since $r_{i}$ is defined as the distance between the agent and the static object $i$; and considering the state of such object as $X_{i}^{O} = [x^i_{o}\quad y^i_{o}\quad 0 \quad0]^T$. It is possible to write the term $ r_{i}$ as follows: 

$$
 r_{i}=\sqrt{\lVert    Z - HX^{O}_{i}  \rVert}
$$

Where $Z$ represents the measurement of the agent position. Since $X^{O}_{i}$ remains constant for fixed objects, the distance $r_{i}$ only depends on the position of the agent.

It is possible to obtain the velocity field generated by an object $i$ all over a two-dimensional plane by looking at the behavior of agents in each point in the space $\mathbb{R}^2$ as it is described by the following equation:  

\begin{equation} \label{eq4aaf}
\upsilon_{i}(r_i)\hat{r} =V^i_{Z \in \mathbb{R}^2} \hat{V}
\end{equation}

Where $V^i_{Z \in \mathbb{R}^2}$ represents the magnitude of the velocity field measured from a set of measurements that cover all the space $\mathbb{R}^2$ and $\hat{V}$ represents the direction of such velocities in every point. 

Since in real cases there is only sparse information of the agents' motion, it is necessary to consider a method to generalize each velocity field from the measured information. This step is discussed in section \ref{sssec:ANNApproach}.

The objective of this work is to build a set of dynamical models through the estimation of velocity fields  $V^i_{Z \in \mathbb{R}^2}$ related to static objects. Such models can be used as future estimators to hierarchically characterize new velocity fields produced by other unknown objects in an environment. 

\subsection{Kalman Filtering Formulation}
\label{sssec:KF Formulation}

It is proposed a hierarchical method where Kalman filters (KFs) are used to track the motion of each agent in order to obtain information concerning unknown static objects inside the environment.

KFs are formulated as more observations related to new objects are obtained. Each unknown object is modeled separately in order to characterize its effects independently. 

Velocity fields produced by static objects are characterized through perturbations in the velocity of agents that are captured by the innovations of KFs.

The effect of an object $i$ is encoded in a new Kalman filter as a control vector $V^i_k$, see equation (\ref{eq4aa}) and (\ref{eq4ab}). Therefore, KFs integrating characteristics of different objects can be obtained and used to estimate posterior information of other objects inside the scene.

This hierarchical characterization with incremental learning of the effects of unknown objects is fundamental when analyzing real scenarios: in this way, it is possible to start with an approximation of velocity fields generated in normal situations, i.e., given by the motion of agents with routine behaviors. Afterwards, when abnormal situations happen, it is possible to find new objects that explain such variations from the normality.

\begin{figure}[htb]
\centering
\centerline{\includegraphics[width=8.3cm]{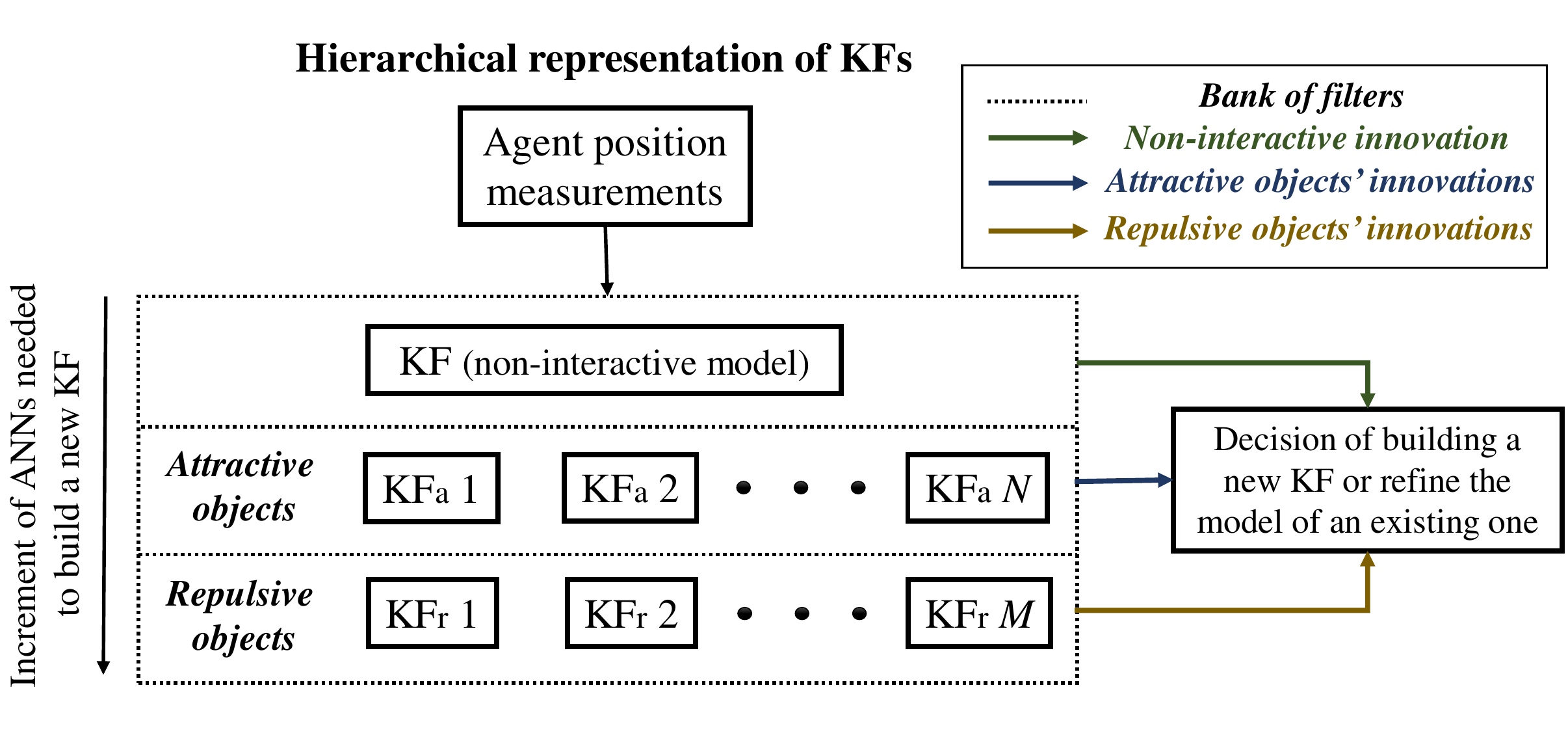}}
\caption{Hierarchical representation of KFs.}
\label{fig:KFs}
\end{figure}

Through the approximation of velocity fields, it is possible to extract information about locations and effects of objects that cause perturbations in the agents' motions. Such information is incrementally added to KF models whose innovation will work as sensors for detecting and characterizing future unknown static objects, see Fig. \ref{fig:KFs}.

A bank of KFs is created from a first filter that uses a non-interactive dynamic model described in equation (\ref{eq4ac}). This consideration allows to start building new KFs that include different dynamical models explaining more complex situations and from which unknown effects in the environment can be learned.

Effects of objects that appear one by one in the scene are added based on equation (\ref{eq4aad}), where each effect is added as a control input vector into a KF formulation.

Regarding the observed information, it is considered the equation (\ref{eq5}), in which $Z_k$ is a measure obtained through an external sensor, such as a real-time locating system, a GPS or a radar. In case a nonlinear sensor is used, an Extended Kalman Filter (EKF) can be used to extend the proposed approach. 

The method described in the present work can be of great help where video surveillance cameras are not available but only information about the localization of individuals can be used.  In several situations, measurements of cameras are not reliable due to problems such as noise in images, partial or full occlusion, scene illumination changes, etc. \cite{Yilmaz2006}. In those cases, even partial information about location of moving subjects (e.g., acquired from smartphones) could characterize possible abnormalities in an environment. 

The proposed approach is useful in diverse scenarios, as detection of crashes and wrong parked objects in streets \cite{Lan2015} or for fire detection with surveillance purposes \cite{Foggia2015}. In order to characterize static objects with few samples, the fitting process described in the next section is proposed.

\subsection{Artificial Neural Networks for Velocity Field Fitting}
\label{sssec:ANNApproach}

An Artificial Neural Network (ANN) is used for approximating each velocity field generated by an unknown object. The inputs of each ANN are the observed agent's two-dimensional coordinates $Z$ and its outputs are the agent's velocity components in such position $V_Z$ obtained through dividing the KF innovation by the sampling time, see equation (\ref{eq4ae}).

\begin{figure}[htb]
\centering
\centerline{\includegraphics[width=8cm]{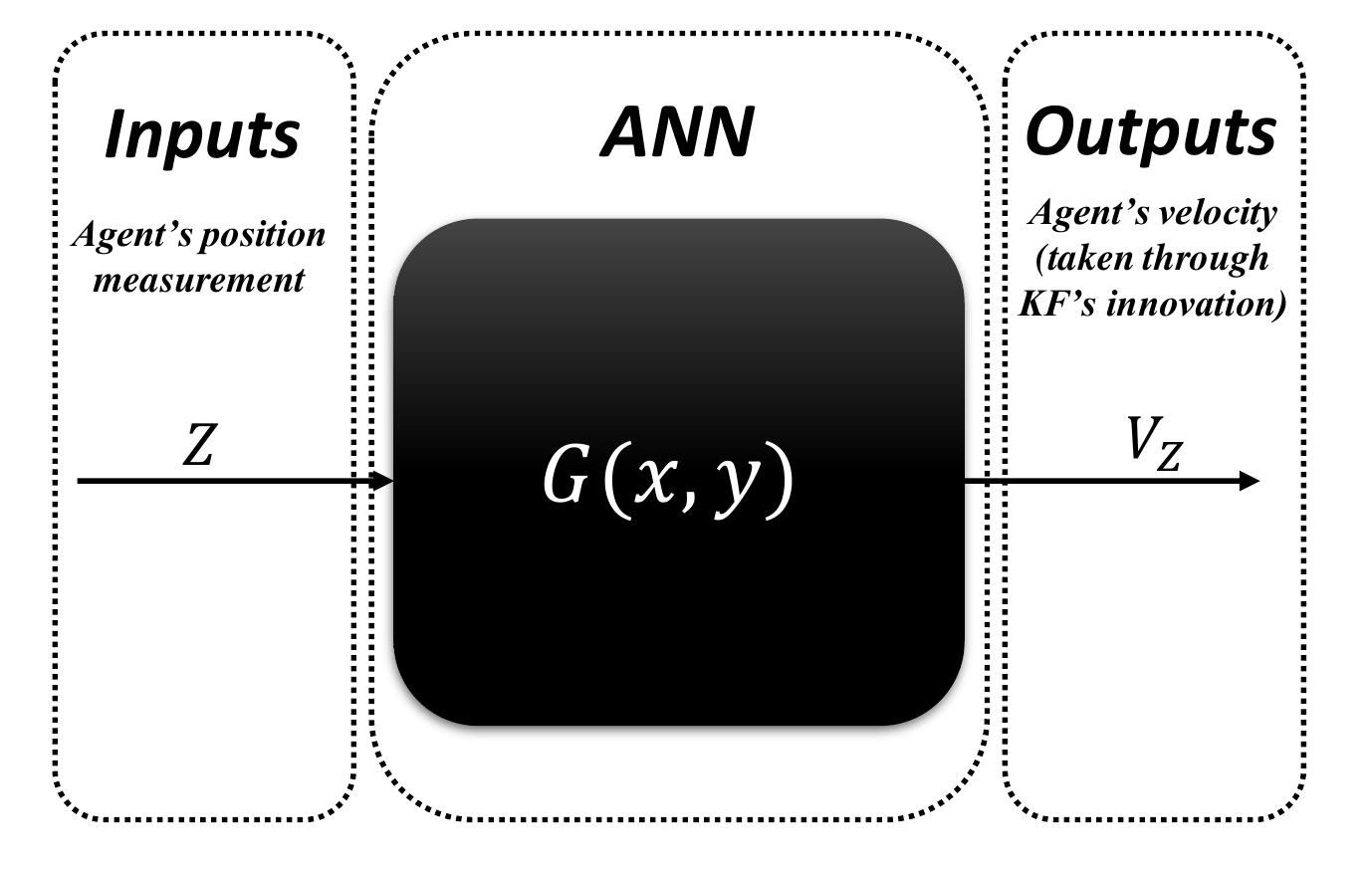}}
\caption{General scheme of the proposed ANN.}
\label{fig:Second}
\end{figure}

The final objective of the fitting process is to find a nonparametric function $G(x,y)$ that relates the coordinates over the whole two-dimensional environment with the velocity field generated by an unknown object, see Fig. \ref{fig:Second}.

The function $G(Z \in \mathbb{R}^2)$ is an approximation of the complete velocity field in $\mathbb{R}^2$. In this way, by taking equation (\ref{eq4aaf}) as reference, it is possible to obtain the relations shown in equation (\ref{eq4aag}) for an object $i$.

\begin{equation} \label{eq4aag}
\upsilon_{i}(r_i)\hat{r} =V^i_{Z \in \mathbb{R}^2} \hat{V} \approx G^i(Z \in \mathbb{R}^2)\hat{g}
\end{equation}

Where $\hat{g}$ represents the direction of the velocity produced by the approximation with an ANN.   

For each static object inside the environment, it has to be considered an ANN that models its behavior. The architecture of each ANN consists of one hidden layer of $10$ neurons that uses the hyperbolic tangent sigmoid as neural transfer function and it is trained based on the Levenberg-Marquardt backpropagation training function.

In this work, for the characterization of the velocity field generated by each object, it was considered a total of $100$ agent trajectories that are affected by it. The starting points of agents are chosen randomly to obtain spread information inside the scene. Each ANN is trained taking into account the inputs and outputs shown in Fig.\ref{fig:Second}. 

\subsection{Extraction of Information Related to Unknown Objects}
\label{sssec:InfoExtract}

Given a trained ANN that approximates the velocity field $\upsilon(r)\hat{r}$ of an unknown object all over a two-dimensional plane, it is possible to infer its center of the force by considering the divergence of the vector field $\nabla\cdot\upsilon(r)\hat{r}$. 

Assuming that objects produce central force fields, the algorithm \ref{Algo1a} is considered for inferring their center of force and nature, i.e., attractive or repulsive.   

\begin{algorithm}
\label{Algo1a}
 \KwData{-Trained ANN according with Fig \ref{fig:Second}. \\
- Discretization rate: $step$, for environment division.}
 \KwResult{Position of the object $objx$ and $objy$.\\
-Nature of the object: $class$.}
$\upsilon\gets$ Vector field produced by an ANN in an environment divided using a stepsize $step$ in $x$ and $y$\\
$div=divergence(\upsilon)$\tcp*{Divergence of the vector field generated by ANN.}
$absdiv=abs(div)$\tcp*{Absolute value of the divergence of the velocity field.}
$[objx,objy]=max(absdiv)$\tcp*{Extraction of coordinates with maximum value.}
$signdiv=sign(div(objx,objy))$\tcp*{Evaluation of sign in particular point of $div$.}
\If{$signdiv< 0$}{
$class= ``Attractive"$\tcp*{Object classified as attractive.} 
\Else{
$class= ``Repulsive"$\tcp*{Object classified as repulsive.}
}
}
 \caption{Extraction of the object location and nature.}
\end{algorithm}

As it can be seen from the algorithm \ref{Algo1a}, by having the ANN approximation of the velocity field $G(Z \in \mathbb{R}^2)\hat{g}$, it is possible to extract the position and nature of an object that exerts a central force field, this knowledge is useful for understanding environment characteristics and to establish relations of causality between static objects and mobile agents in the scene.

\section{RESULTS}
\label{sec:results}

\subsection{Experimental setup}
\label{sssec:ExpSetup}

For demonstration purposes, an environment is defined in a two-dimensional space whose measurements go from $-20$ to $20$ in both axes. The size of the environment does not have any consequence in the performance of the proposed method in terms of functionality. Although, for better resolution in the results it is necessary to divide the environment in several parts in order to evaluate the ANNs with more points. In this sense, it is an advantage to represent objects in small environments.

Compact scenes also present the advantage of needing few trajectories to estimate the velocity field in the entire plane, i.e., the smaller the environment, the less samples are needed to obtain a good representation of it even with sparse data.

Two static attractive objects with different centers of force are placed inside the environment. They will be called first attractor, labeled as $a_{1}$ and second attractor, labeled as $a_{2}$. Their respective centers of force are $(0,15)$ and $(-10,10)$; and their velocity fields are $\upsilon_{a_{1}}$ and $\upsilon_{a_{2}}$ whose formulations are shown in equations (\ref{eq1}) and (\ref{eq2}), respectively.

\begin{equation} \label{eq1}
\upsilon_{a_1} (d_{a_{1}}) =
\begin{cases}  
      	\sqrt{d_{a_1}}/b_{1} & 0\leq d_{a_1}\leq c_{1} \\
	\sqrt{c_{1}}/b_{1} &  c_{1}< d_{a_1} \le f_{1}\\
      	0 & f_{1}> d_{a_1}
   \end{cases}
\end{equation}

\begin{equation} \label{eq2}
\upsilon_{a_{2}} (d_{a_{2}}) =
\begin{cases}  
      	b_{2} e^{-(d_{a_{2}}-c_{2})^2/\alpha_{1}} & 0\leq d_{a_{2}}\leq c_{2} \\
	b_{2} &  c_{2}< d_{a_2} \leq f_{2}\\
      	0 & f_{2}> d_{a_2}
   \end{cases}
\end{equation}

Where $d_{a_{1}}$ and $d_{a_{2}}$ represent the distance between an agent and the center of force of the respective attractor. $b_{1}$ and $b_{2}$  are constants that control the amplitude of the velocity field. $c{1}$ and $c{2}$ are constants related to the distance from the center of force in which agents start decreasing their velocity while they are approaching to their destinations. $f_{1}$ and $f_{2}$ are constants related to the distance from the center of force in which the attractor does not have any influence. $\alpha_{1}$ represents the way in which the second attractor makes agents decrease their velocities as they approach to it. 

Moving agents go towards either of two attractors. The starting point of agents is randomly selected, assuming that agents can appear from whatever point in the environment. Once an agent reaches its destination is removed from the scene and no more information is received from it.

A time interval $\Delta k=1$ is considered as the sampling time that a locating system requires to acquire information about the agent positions inside the environment. Each time instant $k$, a Gaussian noise is applied to the coordinate positions $(x, y)$ of each agent in order to simulate perturbations caused by external factors different from the velocity fields produced by objects in the environment. 

Subsequently, a repulsive object is introduced in the scene, its center of force is placed in $(0,-5)$ and equation $ (\ref{eq3})$ is used for computing its velocity field. 

\begin{equation}\label{eq3}
\upsilon_r(d_r)=b_{3} e^{(d_r)^2/\alpha_{2} } 
\end{equation}

Similar to the attractors described previously, $d_r$ represents the distance between the agent and the repulsive object. $b_{3}$ is a constant that affects the magnitude of the velocity field and $\alpha_{2}$ is a constant related to the way in which the velocity field increases when an agent is approaching towards the center force of the object. 

The constants considered in the section \ref{sssec:ExpSetup} for calculation purposes are presented below.

\begin{gather*}
b_{1}=2   \textbf{;}  \quad   b{2}=1.1   \textbf{;}  \quad  b_{3}=0.8 \\
c_{1}=4   \textbf{;}  \quad   c{2}=8 \\
f_{1}=80   \textbf{;}  \quad   f{2}=80 \\
\alpha_{1}=50 \textbf{;}  \quad \alpha_{2}=1000
\end{gather*}

\begin{figure}[htb]
\centering
\centerline{\includegraphics[width=8.7cm]{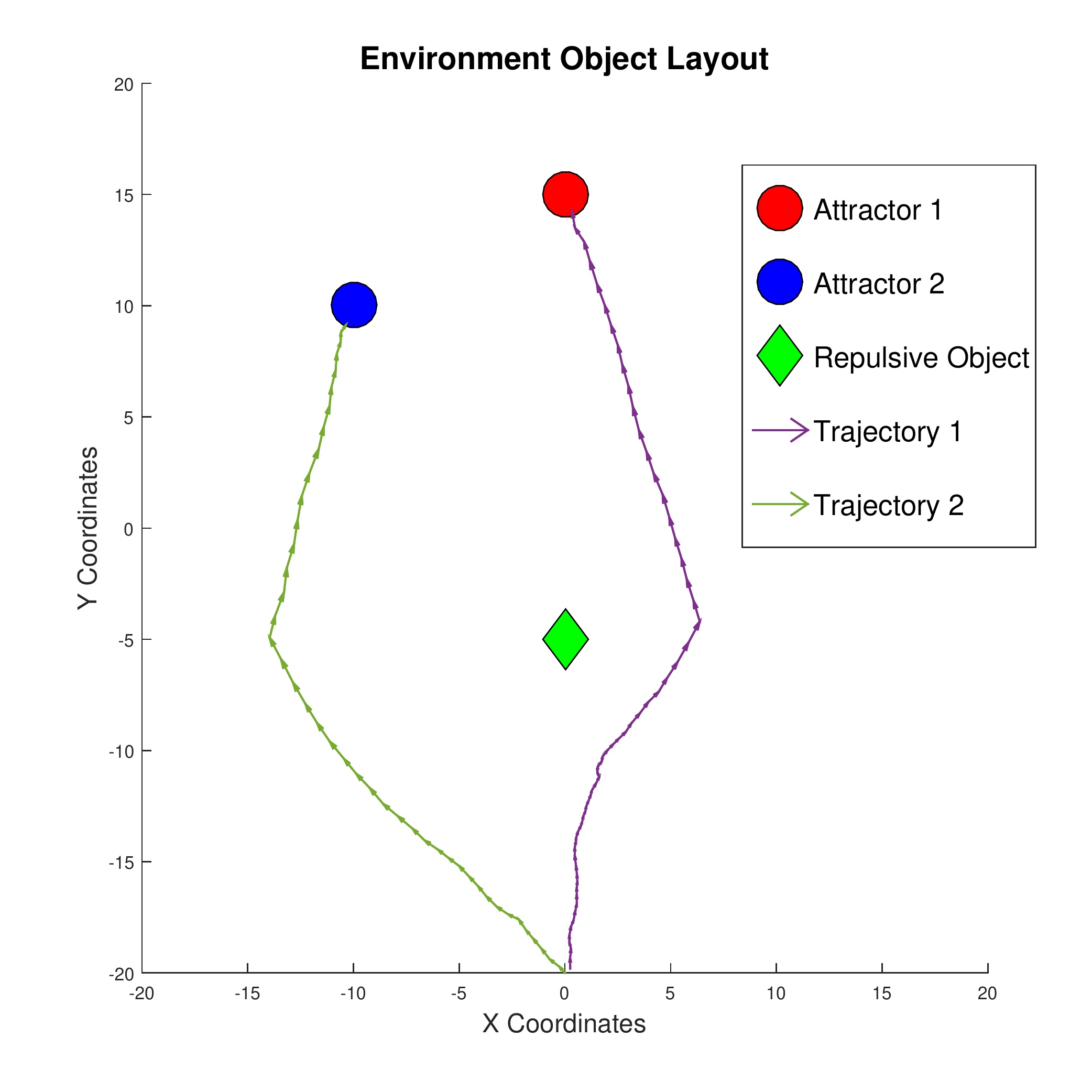}}
\caption{Trajectories of agents with different destinations and same starting point, both are influenced by a repulsive object placed in the middle of their paths.}
\label{fig:First}
\end{figure}

The velocity field of each object is taken as an unknown parameter to estimate, the theoretical expressions shown in equations (\ref{eq1}), (\ref{eq2}) and (\ref{eq3}) are used to measure the performance of the proposed method. Accordingly, the particular situation that will be taken into consideration to evaluate the proposed method is depicted in the Fig. \ref{fig:First}.

Next section shows the results obtained by applying the methodology explained in section \ref{sec:format} over the scenario described in the present section. Velocity fields are obtained by applying the proposed method and are compared with the theoretical ones.  

\subsection{Evaluation of the method}
\label{sssec:ExpSetup}

\begin{figure}[!htb]
  \begin{minipage}[b]{0.48\columnwidth}  
    \centering
    \includegraphics[width=1.05\linewidth]{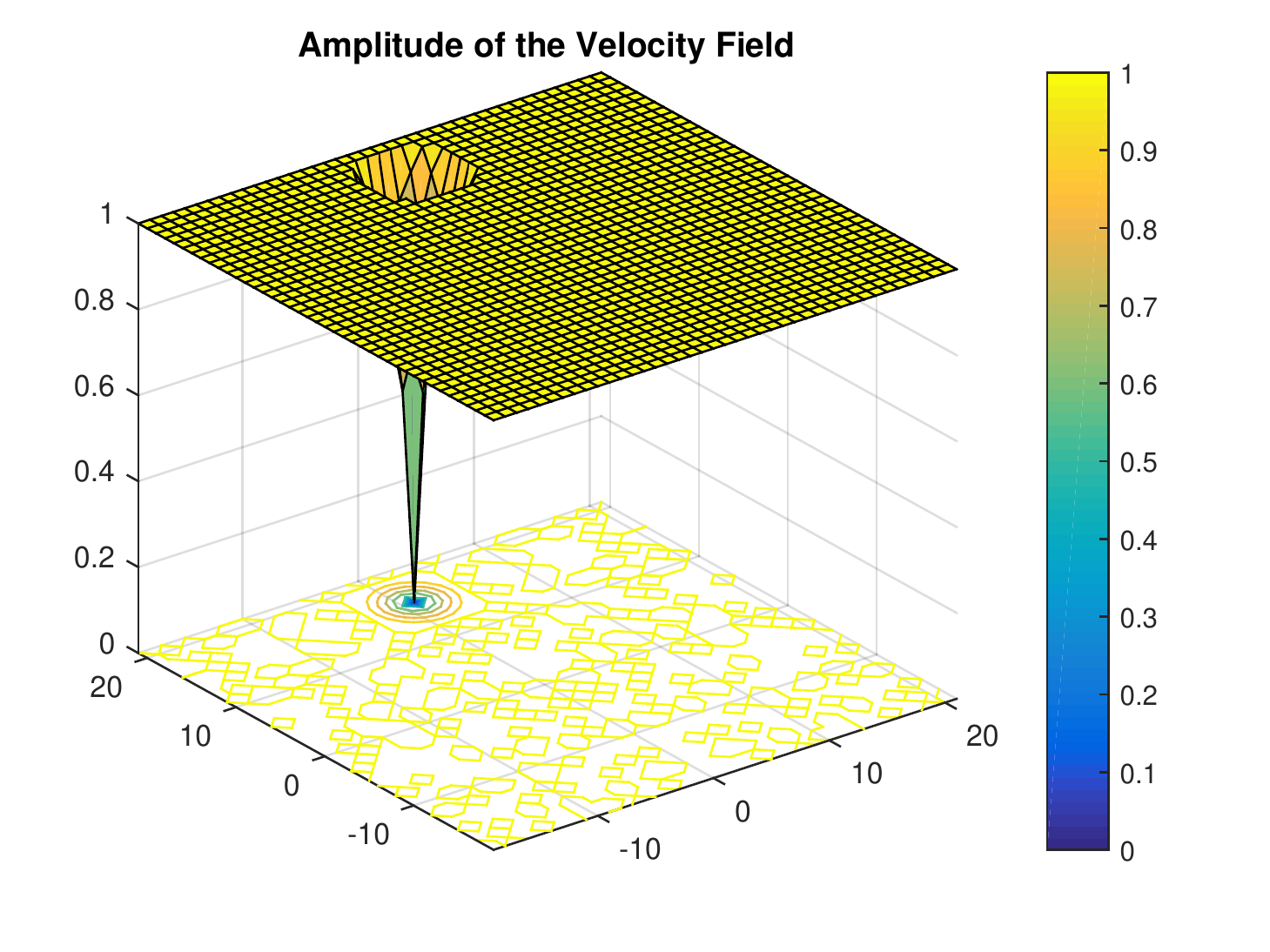} 
    \caption{Theoretical velocity field for first attractor} 
     \label{fig:Third}
    \vspace{3ex}
  \end{minipage}%%
\hfill
  \begin{minipage}[b]{0.48\columnwidth}
    \centering
    \includegraphics[width=1.05\linewidth]{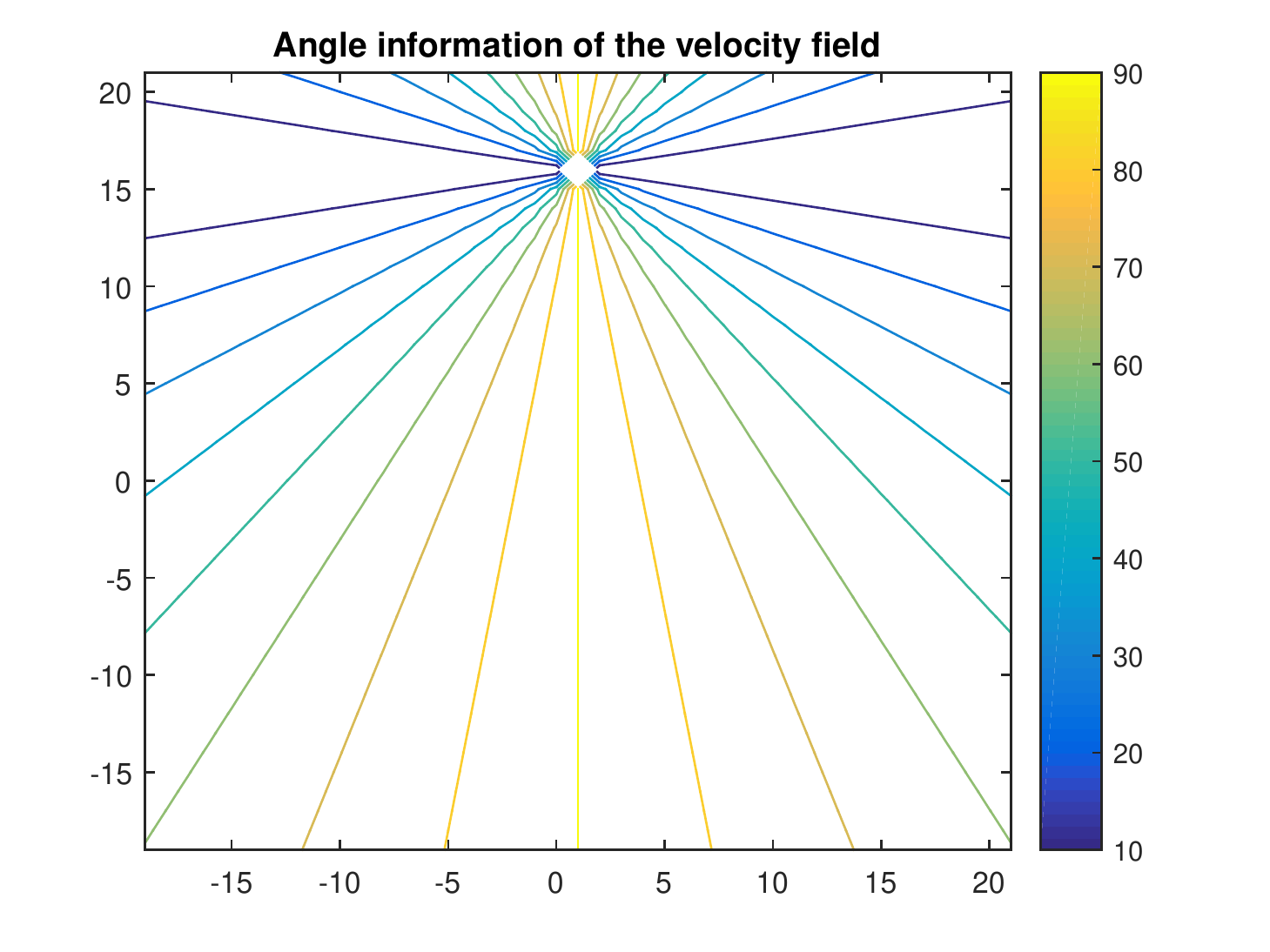} 
    \caption{Theoretical orientation for first attractor field} 
    \label{fig:Fourth}
    \vspace{3ex}
  \end{minipage} 
\hfill
  \begin{minipage}[b]{0.48\columnwidth}
    \centering
    \includegraphics[width=1.05\linewidth]{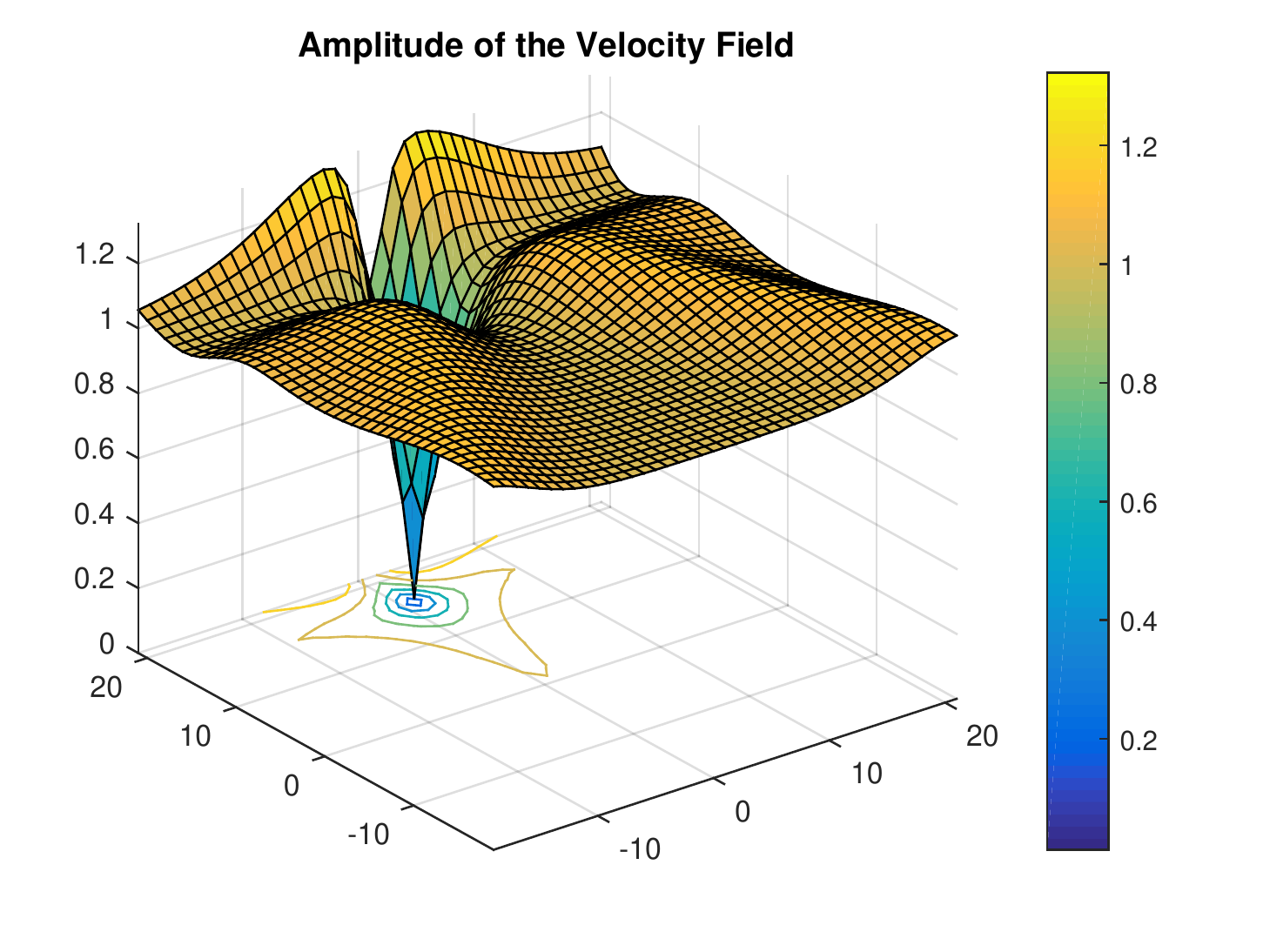} 
    \caption{ANN approx. velocity field for first attractor} 
    \label{fig:Fifth}
    \vspace{3ex}
  \end{minipage}%% 
\hfill
  \begin{minipage}[b]{0.48\columnwidth}
    \centering
    \includegraphics[width=1.05\linewidth]{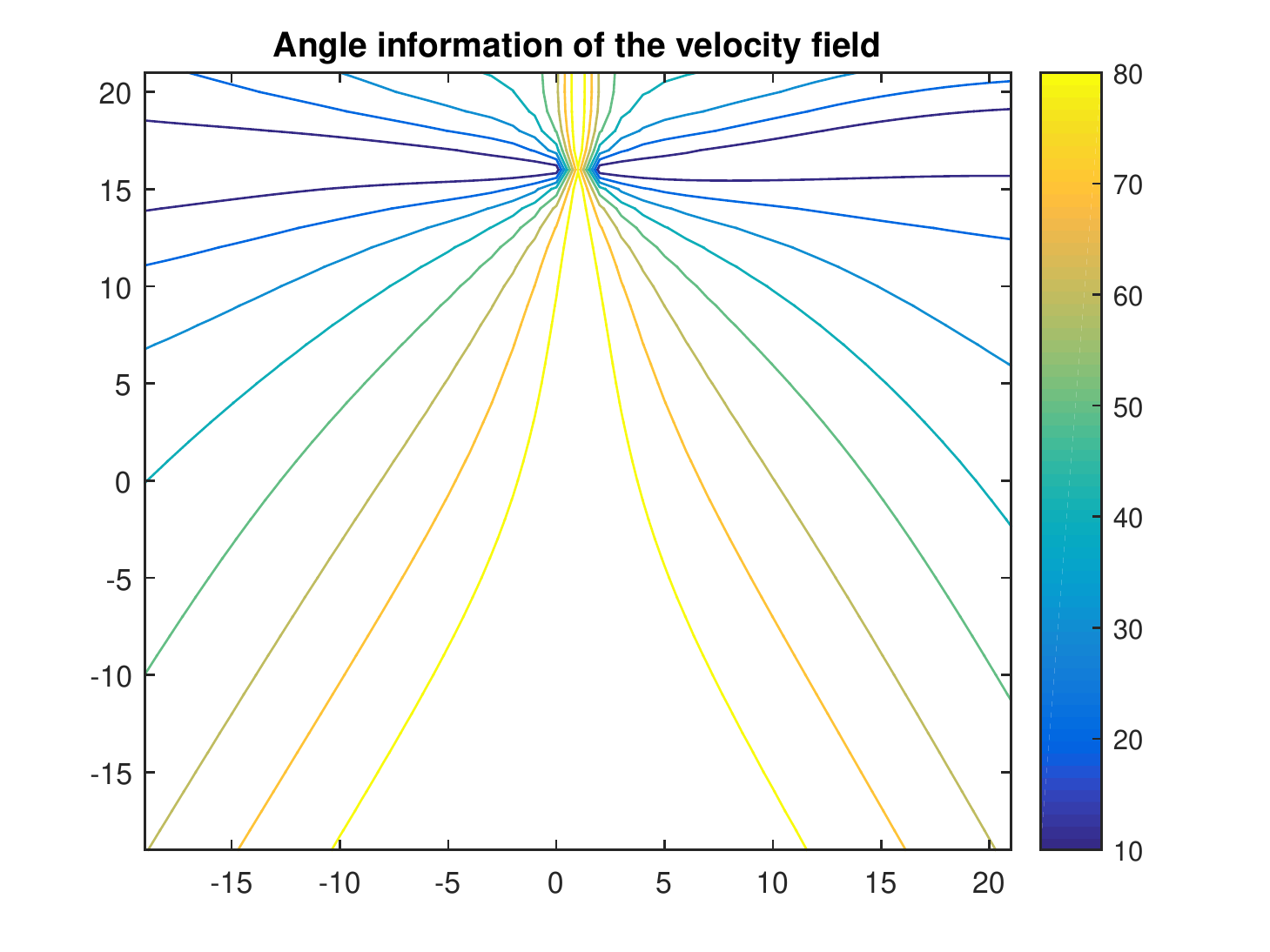} 
    \caption{ANN approx. orientation for first attractor field} 
    \label{fig:Sixth}
    \vspace{3ex}
  \end{minipage} 
\end{figure}

For the first attractor, by taking into consideration its theoretical velocity field described in the equation (\ref{eq1}), it is possible to obtain the graphic of Fig. \ref{fig:Third}. Afterwards, by considering $\theta$ as the angle of the velocity field evaluated in a particular point, it is possible to calculate $|\theta|$, whose behavior is plotted in Fig. \ref{fig:Fourth}. 

By applying the Euclidean norm on the velocity field generated by the ANN approximation for the first attractor, it is possible to obtain the graph of Fig. \ref{fig:Fifth}. By calculating the absolute value of the angle formed by the velocity generated by the ANN, it is possible to obtain Fig. \ref{fig:Sixth}.

By comparing Fig. \ref{fig:Third} and Fig. \ref{fig:Fourth} with Fig. \ref{fig:Fifth} and Fig. \ref{fig:Sixth} respectively, it can be seen that the proposed formulation provides a reliable approximation to the real velocity field produced by an object inside the scene.

\begin{figure}[!htb]
  \begin{minipage}[b]{0.48\columnwidth}  
    \centering
    \includegraphics[width=1.05\linewidth]{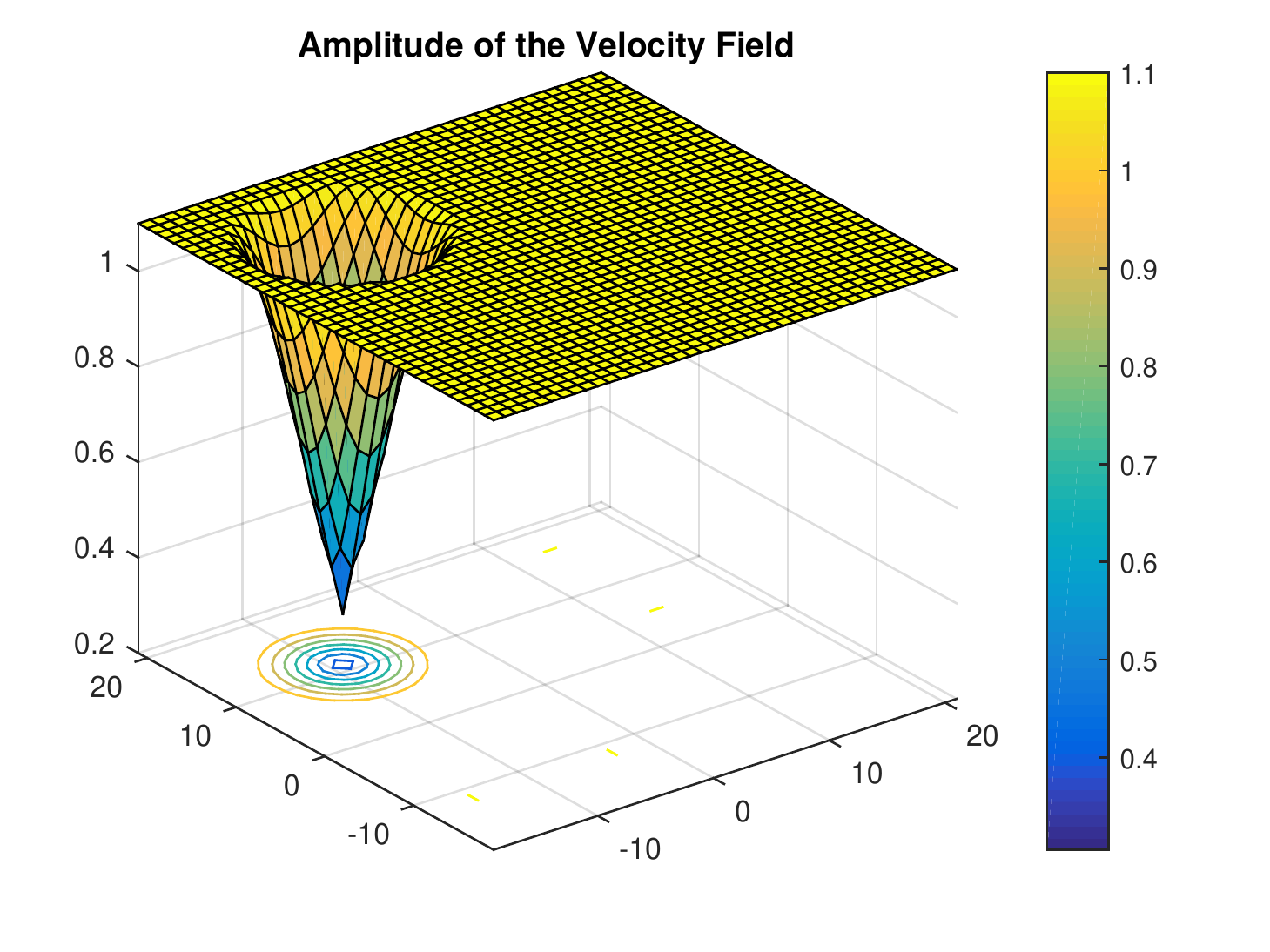} 
    \caption{Theoretical velocity field for 2nd attractor} 
     \label{fig:Seventh}
    \vspace{3ex}
  \end{minipage}%%
\hfill
  \begin{minipage}[b]{0.48\columnwidth}
    \centering
    \includegraphics[width=1.05\linewidth]{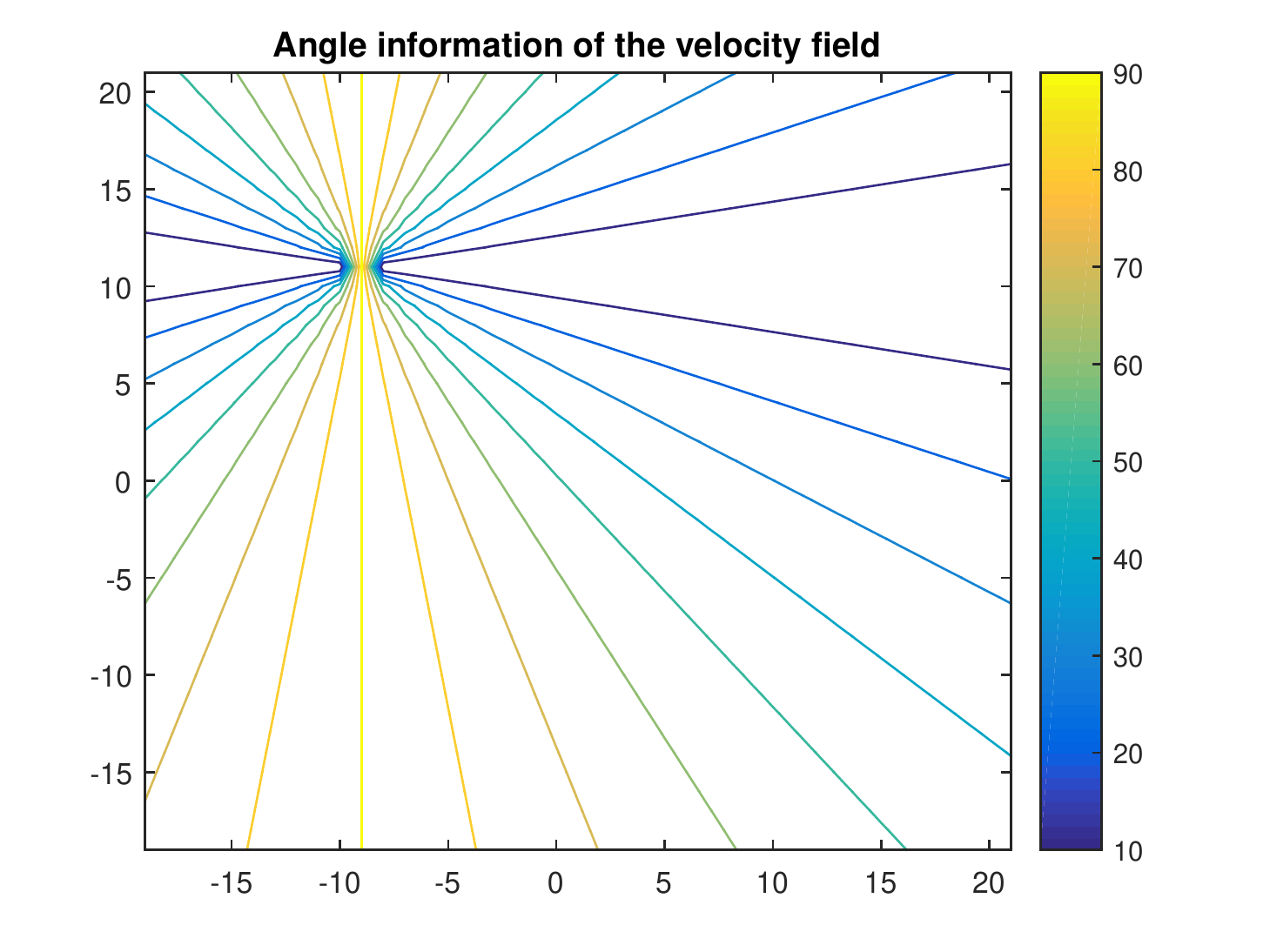} 
    \caption{Theoretical orientation for 2nd attractor field} 
    \label{fig:Eighth}
    \vspace{3ex}
  \end{minipage} 
\hfill
  \begin{minipage}[b]{0.48\columnwidth}
    \centering
    \includegraphics[width=1.05\linewidth]{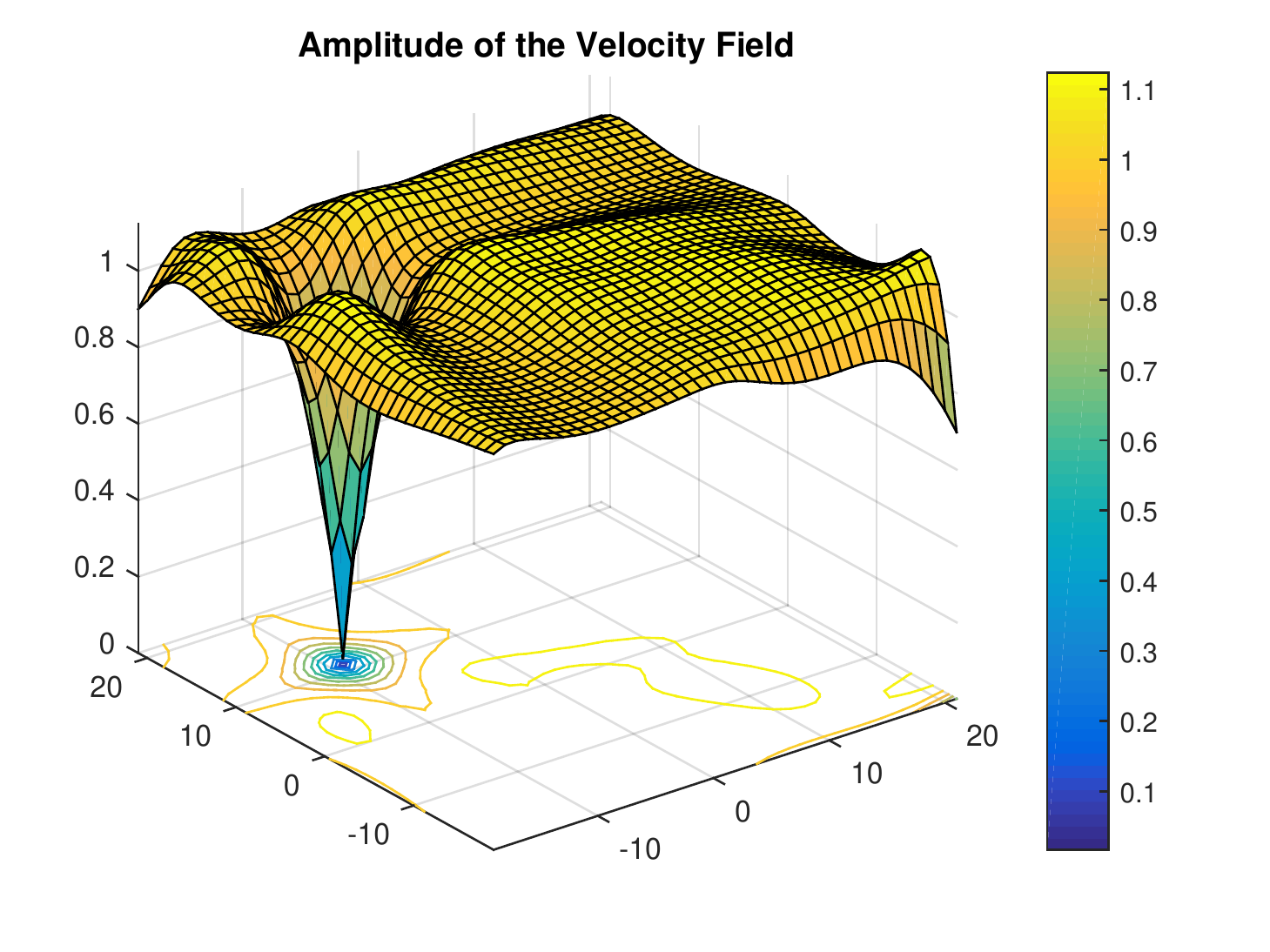} 
    \caption{ANN approx. velocity field for 2nd attractor} 
    \label{fig:Ninth}
    \vspace{3ex}
  \end{minipage}%% 
\hfill
  \begin{minipage}[b]{0.48\columnwidth}
    \centering
    \includegraphics[width=1.05\linewidth]{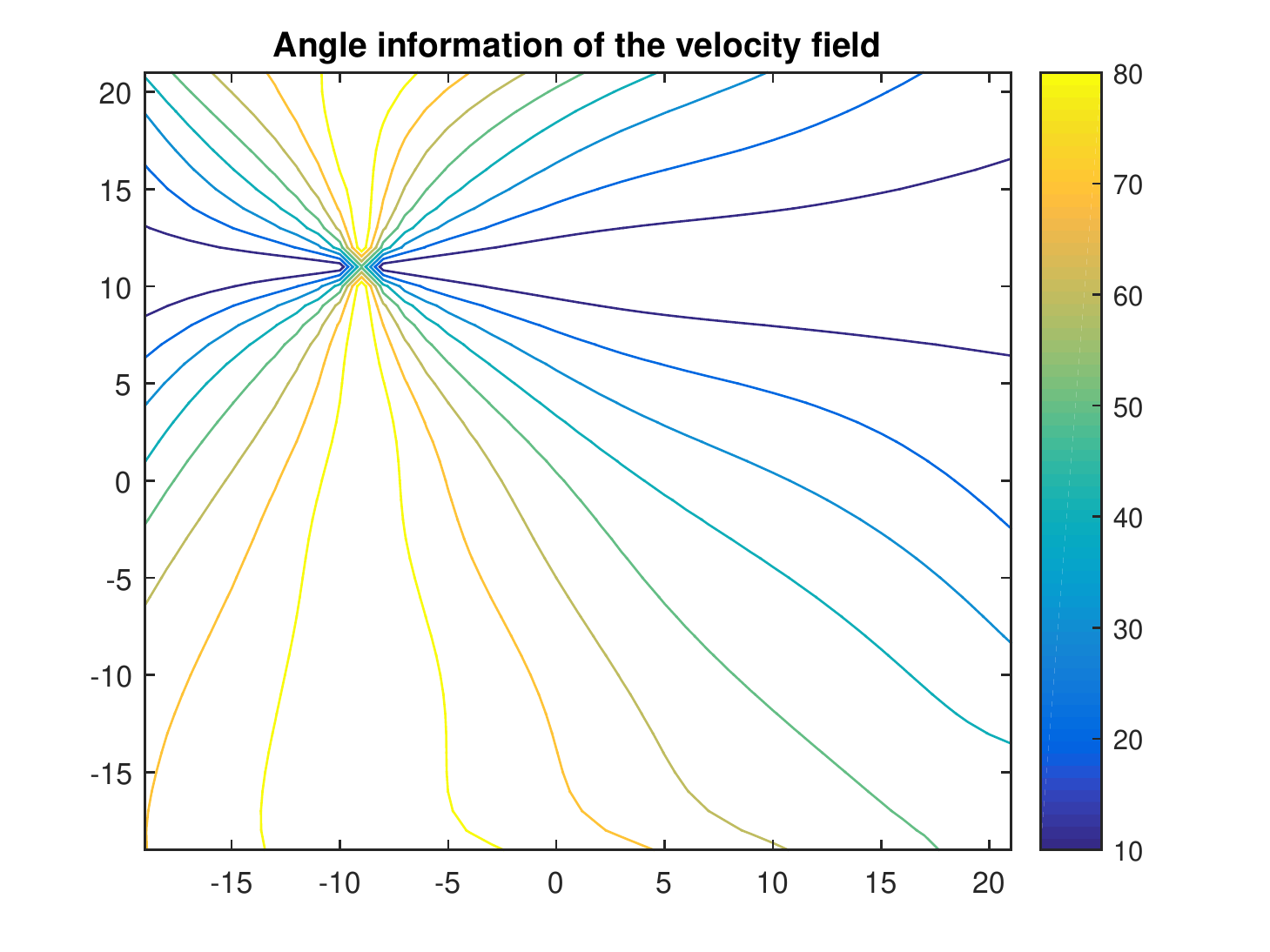} 
    \caption{ANN approx. orientation for 2nd attractor field} 
    \label{fig:Tenth}
    \vspace{3ex}
  \end{minipage} 
\end{figure}

Similarly, the process described for the first attractor was repeated for the other attractive object as well. The ground truth of the velocity field produced by the second attractor and its modulus of the angle are depicted in Fig. \ref{fig:Seventh} and Fig. \ref{fig:Eighth}, respectively. The corresponding approximations by using the proposed method are shown in Fig. \ref{fig:Ninth} and Fig. \ref{fig:Tenth}. 

\begin{figure}[!htb]
  \begin{minipage}[b]{0.48\columnwidth}  
    \centering
    \includegraphics[width=1.05\linewidth]{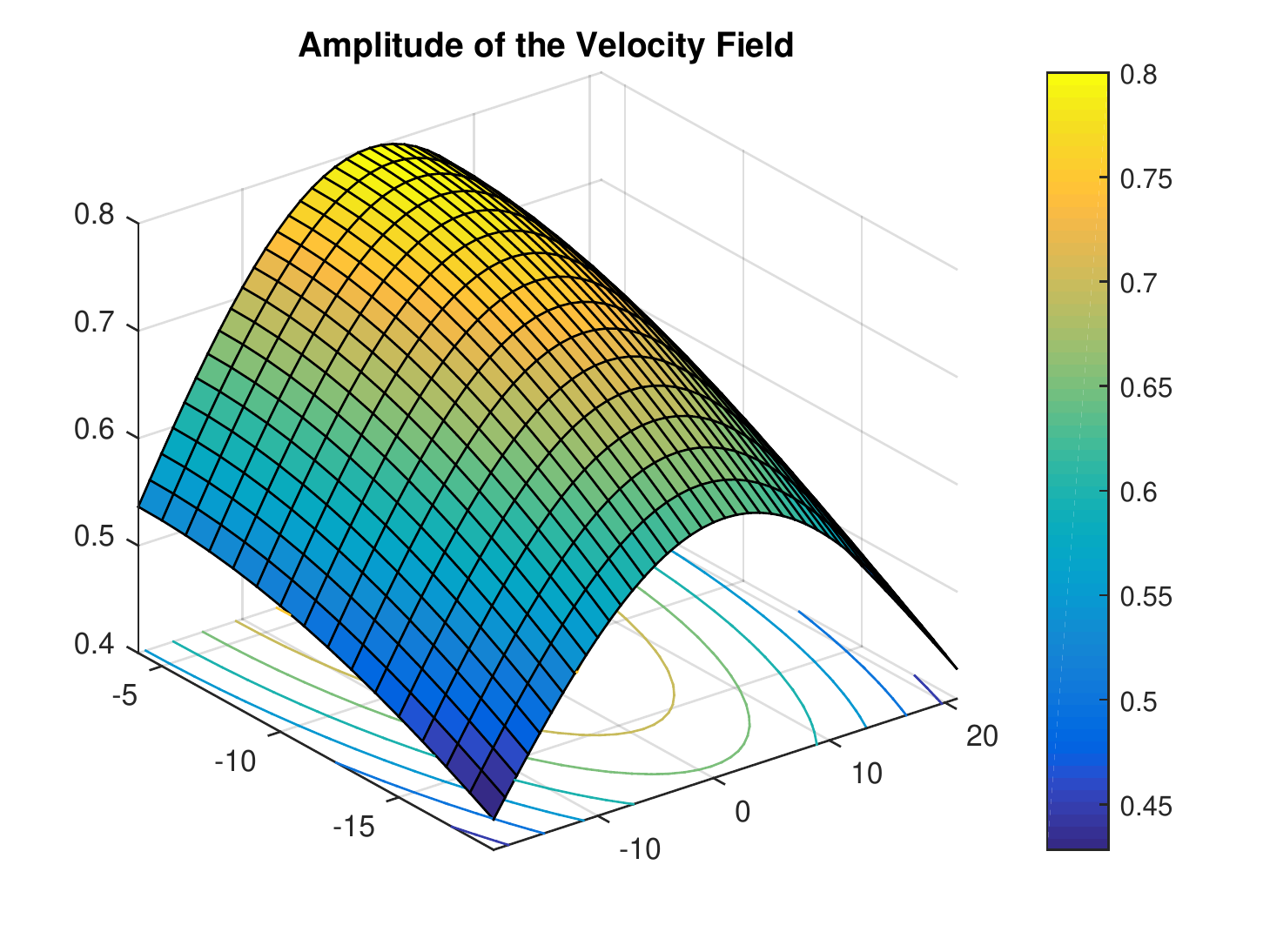} 
    \caption{Theoretical velocity field for the repeller} 
     \label{fig:Eleventh}
    \vspace{3ex}
  \end{minipage}%%
\hfill
  \begin{minipage}[b]{0.48\columnwidth}
    \centering
    \includegraphics[width=1.05\linewidth]{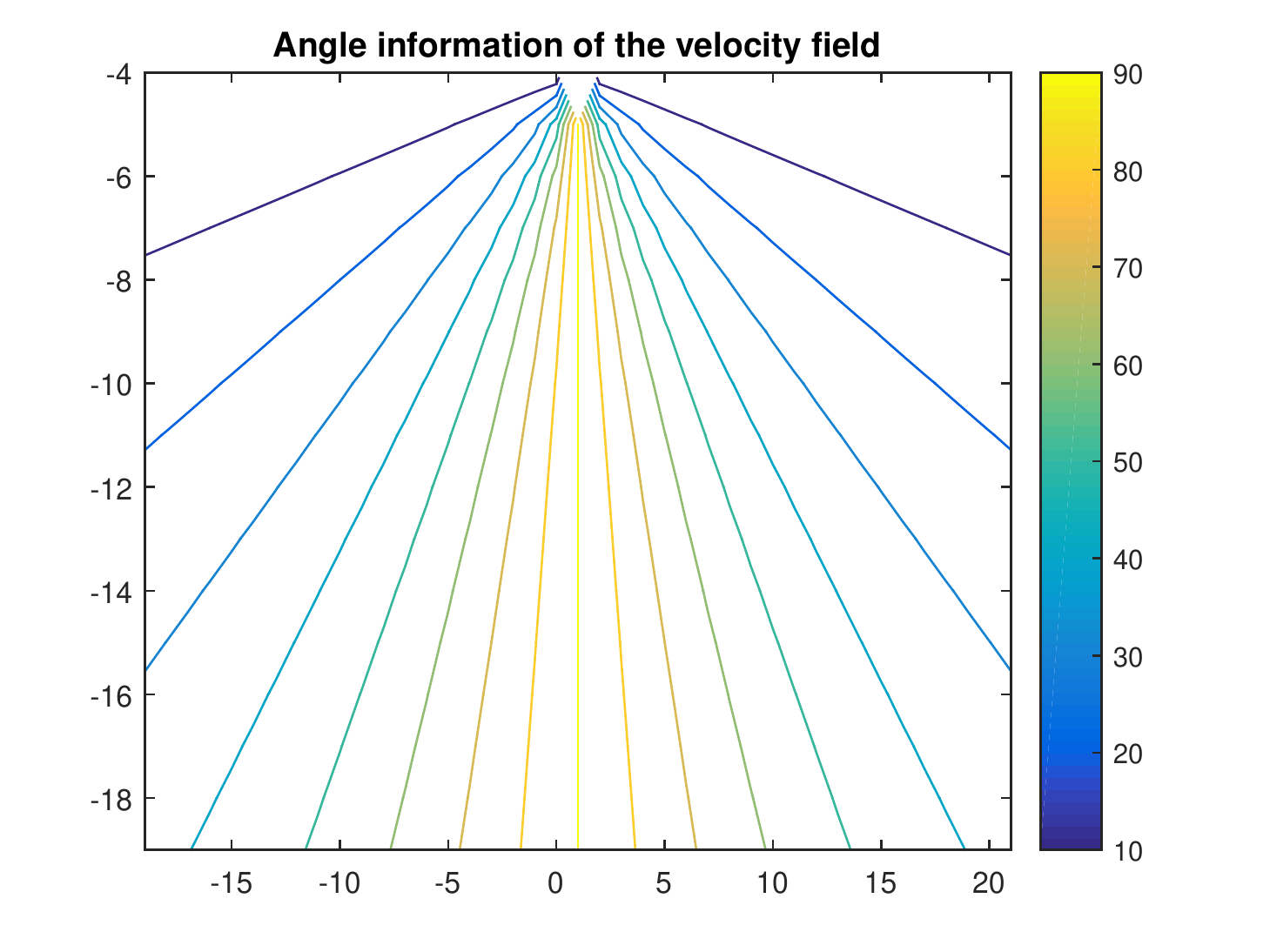} 
    \caption{Theoretical orientation for the repeller field} 
    \label{fig:Twelveth}
    \vspace{3ex}
  \end{minipage} 
\hfill
  \begin{minipage}[b]{0.48\columnwidth}
    \centering
    \includegraphics[width=1.05\linewidth]{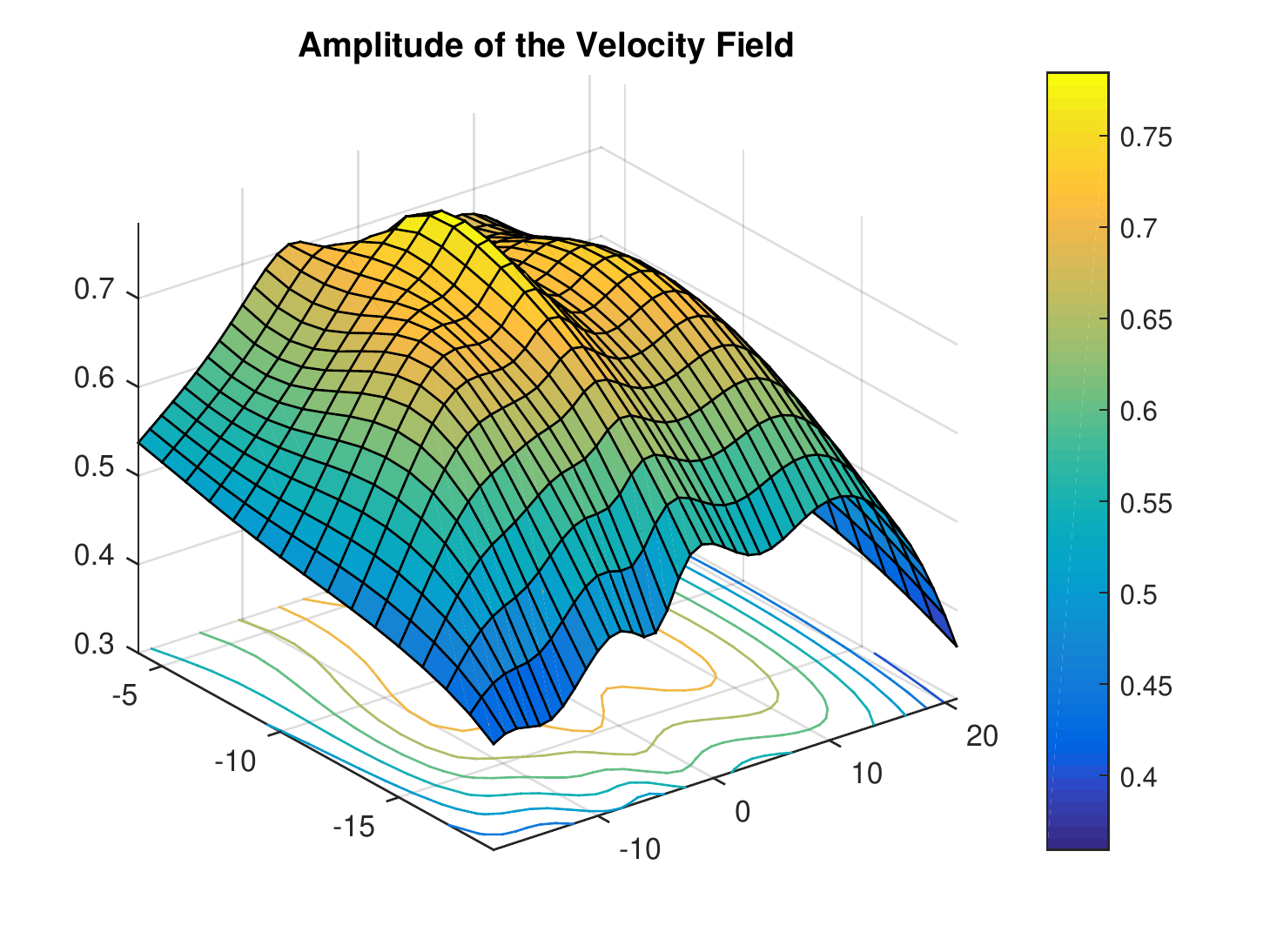} 
    \caption{ANN approx. velocity field for the repeller} 
    \label{fig:thirteenth}
    \vspace{3ex}
  \end{minipage}%% 
\hfill
  \begin{minipage}[b]{0.48\columnwidth}
    \centering
    \includegraphics[width=1.05\linewidth]{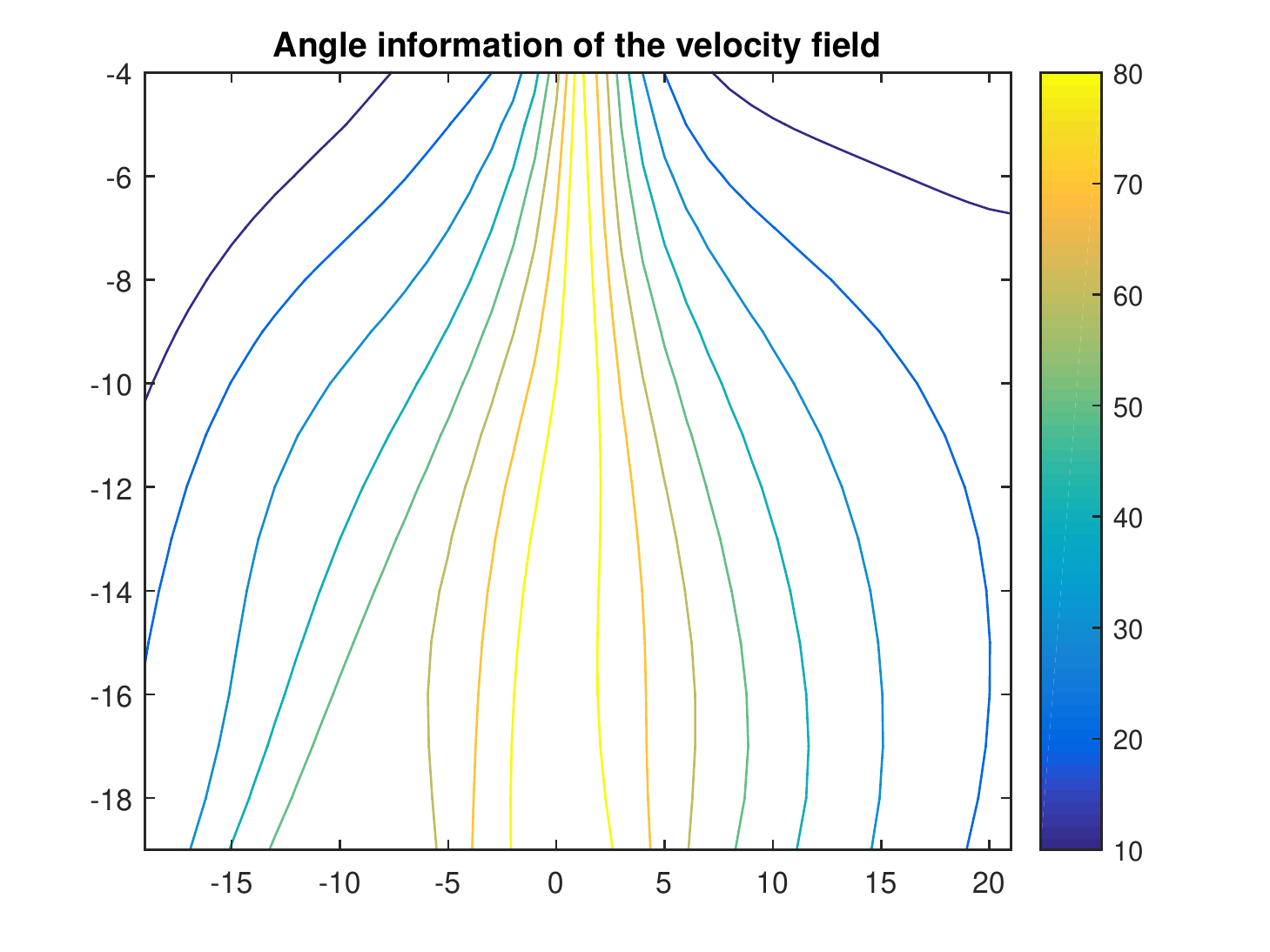} 
    \caption{ANN approx. orientation for the repeller field} 
    \label{fig:Fourteenth}
    \vspace{3ex}
  \end{minipage} 
\end{figure}

For the approximation of the repulsive object, it is important to mention that its characteristics were approximated after having the model for both attractors. Accordingly, the approximated velocity fields generated by both attractors were used as prior knowledge to hierarchically characterize the velocity field of the repulsive object introduced in the scene. The ground truth information of the repulsive object is depicted in Fig. \ref{fig:Eleventh} and Fig. \ref{fig:Twelveth}. The approximation using the proposed method is shown in Fig. \ref{fig:thirteenth} and Fig. \ref{fig:Fourteenth}.

For extracting the position of each object and its nature, i.e., attractive or repulsive, the algorithm \ref{Algo1a} was used with $step = 0.3$. Overall results are shown in Table \ref{table:1}.

\begin{table}[h!]
\caption{Results of localization and nature of objects }
\centering
 \begin{tabular}{||c c c c ||} 
 \hline
 \textbf{Object} & \textbf{Real Position} &\textbf{Estimated Position} & \textbf{Nature} \\ [0.5ex] 
 \hline\hline 
 Attractor1 & (0, 15) & (-0.2, 15.1) & Attractive  \\ [0.5ex] 
 Attractor2 & (-10, 10) & (-10.1, 10) & Attractive  \\
 Repeller & (0, -5) & (0.1, -4.7) & Repulsive \\ [0.5ex] 
 \hline
 \end{tabular}
\label{table:1}
\end{table}

From Table \ref{table:1}, it is possible to see that proposed approach has great accuracy in the localization of the center of forces of unknown static objects. It also recognizes correctly the attractive or repulsive behavior of each object.  

\section{CONCLUSIONS}
\label{sec:typestyle}

A method to estimate velocity fields caused by static objects given the motion of agents was formulated and validated for attractive and repulsive objects. The innovation or measurement residual of a KF formulation was proposed to perform the characterizations of such velocity fields. 

The effect of unknown static objects is represented into a KF formulation by a term that consists in a control input model multiplied by a control vector. The additivity of such terms allows to build different dynamical models with more complete information that describe the motion of agents in a hierarchical way. In this sense, each new object that appears in the scene can be added as part of the control vector into a dynamical formulation. This methodology allows to incrementally learn information about the effects inside an environment by analyzing the movement of agents.

Artificial Neural Networks were used to perform a regression over sparse data in order to generalize velocity fields for the whole environment. Theoretical and practical results are compared, showing that the proposed method allows to properly estimate the effects of unknown static objects in two-dimensional environments.   

For a future work, this approach will be the basis of a system for detecting abnormalities in guarded environments. This method can be also extended for the characterization of dynamical moving objects in which online learning techniques should be explored to characterize information inside the scene as time evolves.

% References should be produced using the bibtex program from suitable
% BiBTeX files (here: strings, refs, manuals). The IEEEbib.bst bibliography
% style file from IEEE produces unsorted bibliography list.
% -------------------------------------------------------------------------

\bibliographystyle{IEEEbib}
\bibliography{refs}

\end{document}